\documentclass[journal]{IEEEtran}      

\usepackage{amsmath}
\usepackage{bm}
\usepackage{array}
\usepackage{enumerate}
\usepackage{mathrsfs}
\usepackage{makeidx}         
\usepackage{graphicx}        
\usepackage{multicol}        
\usepackage{subfigure}
\usepackage{epsfig,amssymb,latexsym}
\usepackage{psfrag}
\usepackage{url}
\usepackage{cite}
\usepackage{epsfig,amssymb,latexsym}
\usepackage{psfrag}
\usepackage[ruled,vlined]{algorithm2e}
\usepackage{algorithm2e}

\makeatletter

\renewcommand*\env@matrix[1][\arraystretch]{%
  \edef\arraystretch{#1}%
  \hskip -\arraycolsep
  \let\@ifnextchar\new@ifnextchar
  \array{*\c@MaxMatrixCols c}}
\makeatother

\newcolumntype{L}[1]{>{\raggedright\let\newline\\\arraybackslash\hspace{0pt}}m{#1}}
\newcolumntype{C}[1]{>{\centering\let\newline\\\arraybackslash\hspace{0pt}}m{#1}}
\newcolumntype{R}[1]{>{\raggedleft\let\newline\\\arraybackslash\hspace{0pt}}m{#1}}

\newcommand{\bi}{\begin{itemize}}
\newcommand{\ei}{\end{itemize}}
\newcommand{\be}{\begin{equation}}
\newcommand{\ben}{\begin{equation*}}
\newcommand{\ee}{\end{equation}}
\newcommand{\een}{\end{equation*}}
\newcommand{\bd}{\begin{displaymath}}
\newcommand{\ed}{\end{displaymath}}
\newcommand{\ba}{\begin{align}}
\newcommand{\ea}{\end{align}}
\newcommand{\bc}{\begin{center}}
\newcommand{\ec}{\end{center}}

\newcommand{\diag}{\operatorname{diag}}

\title{\LARGE \bf
Indirect Shared Control of Highly Automated Vehicles for Cooperative Driving between Driver and Automation}
\author{Renjie Li, Yanan Li, Shengbo Eben Li, Etienne Burdet and Bo Cheng
\thanks{This research is supported by NSF China with U1664263 and 51575293, and National Key R\&D Program in China with 2016YFB0100906. All correspondence should be sent to S. Li (e-mail: lisb04@gmail.com).}
\thanks{R. Li is with the State Key Lab of Automotive Safety and Energy, Tsinghua University, Beijing, China 100084. He is currently visiting the Department of Bioengineering, Imperial College London, UK SW7 2AZ  (e-mail: lirenaxe@gmail.com).}
\thanks{S. Li and B. Cheng are with the State Key Lab of Automotive Safety and Energy, Tsinghua University, Beijing, China 100084 (e-mail: lisb04@gmail.com; chengbo@tsinghua.edu.cn).}
\thanks{Y. Li and E. Burdet are with the Department of Bioengineering, Imperial College London, UK SW7 2AZ (e-mail: yanan.li@imperial.ac.uk; e.burdet@imperial.ac.uk).}
}

\begin{document}

\maketitle
\thispagestyle{empty}
\pagestyle{empty}

\begin{abstract}
It is widely acknowledged that drivers should remain in the control loop of automated vehicles before they completely meet real-world operational conditions. This paper introduces an `indirect shared control' scheme for steer-by-wire vehicles, which allows the vehicle control authority to be continuously shared between the driver and automation through unphysical cooperation. This paper first balances the control objectives of the driver and automation in a weighted summation, and then models the driver's adaptive control behavior using a predictive control approach. The driver adaptation modeling enables off-line evaluations of indirect shared control systems and thus facilitates the design of the assistant controller. Unlike any conventional driver model for manual driving, this model assumes that the driver can learn and incorporate the controller strategy into his internal model for more accurate path following. To satisfy the driving demands in different scenarios, a sliding-window detector is designed to continuously monitor the driver intention and automatically switch the authority weights between the driver and automation. The simulation results illustrate the advantages of considering the driver adaptation in path-following and obstacle-avoidance tasks, and show the effectiveness of indirect shared control for cooperative driving.

\end{abstract}

\section{Introduction}
In recent years, several ambitious projects of automated vehicles (or self-driving cars, driverless cars), e.g., Google's self-driving car and Tesla's Autopilot, have realized rapid progresses towards commercialization. While driverless cars are considered as an effective approach to relieve drivers through advanced sensing and navigation technologies, various factors need to be addressed before their ultimate deployment. These factors include technical requirements, safety issues, ethical dilemma \cite{Bonnefon16Science}, and subsequent harsh government regulations \cite{Schreurs16}. In addition, previous studies have shown that imperfect automation may lead to severe human factor problems such as loss of situation awareness, overreliance, distrust, etc \cite{Kaber04,Ma05IJIE}. These problems are usually considered to be caused by drivers being kept out of the operating loop during autonomous driving \cite{Kaber97}.
$\tau$
In light of the problems brought by imperfect self-driving technology, there is an increasing interest to keep the human driver in the control loop, who may have superior capabilities for handling complicated situations. For instance, in some copilot systems \cite{Anderson14ITHMS, Erlien13IFAC, Shia14TITS}, the human driver assumes control most of the time, while the assistant controller only intervenes if it reckons that the vehicle is at risk. Therefore, the support provided by the automation is limited to a short proportion of the time. By contrast, in shared control \cite{Abbink12CTW}, or cooperative control depending on the context \cite{Flemisch12CTW}, the automation remains in the control loop and provides continuous support to the driver, thus significantly reducing their workload.

The shared control concept is commonly implemented on conventional vehicles equipped with mechanical steering systems as the haptic shared control, in which the driver and assistant controller simultaneously apply a control torque on the steering wheel \cite{Abbink12CTW}. Under such a framework, the driver and automation control the vehicle steering wheel cooperatively through physical interaction. The performance enhancement of drivers in haptic shared control has been broadly reported in literature (see the survey of Petermeijer \emph{et al.} \cite{Petermeijer15TOH}) including experimental evaluations \cite{Mulder12HF} and controller design aspects \cite{Saleh13TITS,Soualmi14CCP}. In haptic shared control, drivers can deny the assistant control torque by stiffening the hands using muscles contraction \cite{Burdet01, Franklin03, Franklin07} thus keeping the final control authority (if the assistant torque is designed not to exceed the human resistance limit). The mechanism of haptic shared control has been studied in other fields such as human-robot interaction \cite{LiY15TOR, LiY16TOR, Takagi17NatureHumanBehaviour} where it is referred to as ``motor interaction" or ``joint motor action" \cite{Jarrasse12PLOS, Jarrasse14SAGE}.

In contrast to mechanical steering systems, steer-by-wire technology allows the mechanical decoupling of the steering wheel and road wheels. With steer-by-wire it is possible for the assistant controller to modulate the driver's commanded steering input by the lower-level steering actuator. Therefore, steer-by-wire vehicles are an ideal platform to implement various forms of shared control, wherein the controller can complement the driver's steering input behind the scenes without introducing physical interaction. In this scheme, the driver can only control the vehicle indirectly through the controller, and his control authority entirely depends on how the automation assimilates his control input. Therefore, such a scheme is called ``indirect shared control''. Compared with haptic shared control, indirect shared control can further minimize the driver effort since the driver is not obligated to provide full steering. Moreover, the removal of direct physical interaction may reduce driver discomfort which has been observed in haptic shared control \cite{Mars14TOH}.

Important questions that need to be addressed in order to establish indirect shared control include: 1) how to design the controller algorithm such that it can respect the driver input while exploiting its own path-tracking ability; 2) how to model the driver's adaptive behavior in presence of a given controller such that the system can be evaluated through off-line simulation. Concerning the first problem, Manabu \emph{et al.} \cite{Manabu06IATSS} proposed a weighted summation method to combine the driver input with the automation's desired input, which was implemented on a steer-by-wire vehicle and validated through field tests. In this work the authority weights are static during driving, which does not allow the driver to gain more control authority in a critical situation that cannot be properly handled by the automation. Concerning the second problem, to our knowledge no previous work has modeled the driver adaptation. Some studies on haptic shared control \cite{Saleh13TITS, Zafeiropoulos14ACC} included a driver model in controller design and simulation, but they did not consider the driver adaptation, which has however been observed \cite{Abbink10AiH, Mars14CSMC}. Na and Cole \cite{Na15THMS} employed game theory to describe the driver-automation co-adaptation, whereas it required the automation to exactly predict the driver's reference path. In fact, numerous studies have documented the adaptation inherent to sensorimotor control, which has been modeled through the formation of an internal inverse or forward model \cite{Burdet13MITpress}. This leads to an intuitive conjecture that the driver adaptation in indirect shared control can be interpreted as the inclusion of the assistant controller in their internal model.

\emph{Contributions}: This paper proposes a novel indirect shared control scheme to maintain the driver-in-the-loop function in highly automated steer-by-wire vehicles, which allows an unphysical cooperation between the driver and automation. The assistant controller follows the weighted summation method as in \cite{Manabu06IATSS} to balance the interests of the driver and automation, whereas the control authorities can be automatically switched with respect to the driver intention. To better assess the performance of indirect shared control, this paper proposes a method to model the driver adaptation in presence of the assistant controller. This method relies on the assumption that the human driver can identify the controller strategy through sensorimotor learning. Cole \emph{et al}. \cite{Cole06VSD} first suggested that driver steering behavior can be properly modeled by model predictive control (MPC), in which the internal model of vehicle dynamics serves as the predictor. Here we develop an MPC model of the driver's adaptive behavior which incorporates the identified controller strategy into the driver's internal model for predictive path tracking. The modeling of the driver adaptation enables off-line evaluations of indirect shared control systems and thus facilitates the design of the assistant controller.

This paper is organized as follows. Section \ref{sec.system_description} describes the framework of indirect shared control including the vehicle model to be used. Section \ref{sec.strategies} formulates the strategies of the controller and the driver based on two fundamental assumptions, and Section \ref{sec.derivation} derives the analytical solution of the driver strategy formulated in Section \ref{sec.strategies}. Section \ref{sec.weight_switching} introduces an automatic weight switching method which can overcome the trade-off brought by static weighting. Section \ref{sec.simulation} validates the proposed driver model and the automatic weight switching by simulation, and Section \ref{sec.conclusion} concludes the paper.

\section{System Description} \label{sec.system_description}
\subsection{General Framework}
\begin{figure}[thpb]
      \centering
      \subfigure[]{
      \includegraphics[scale=0.6]{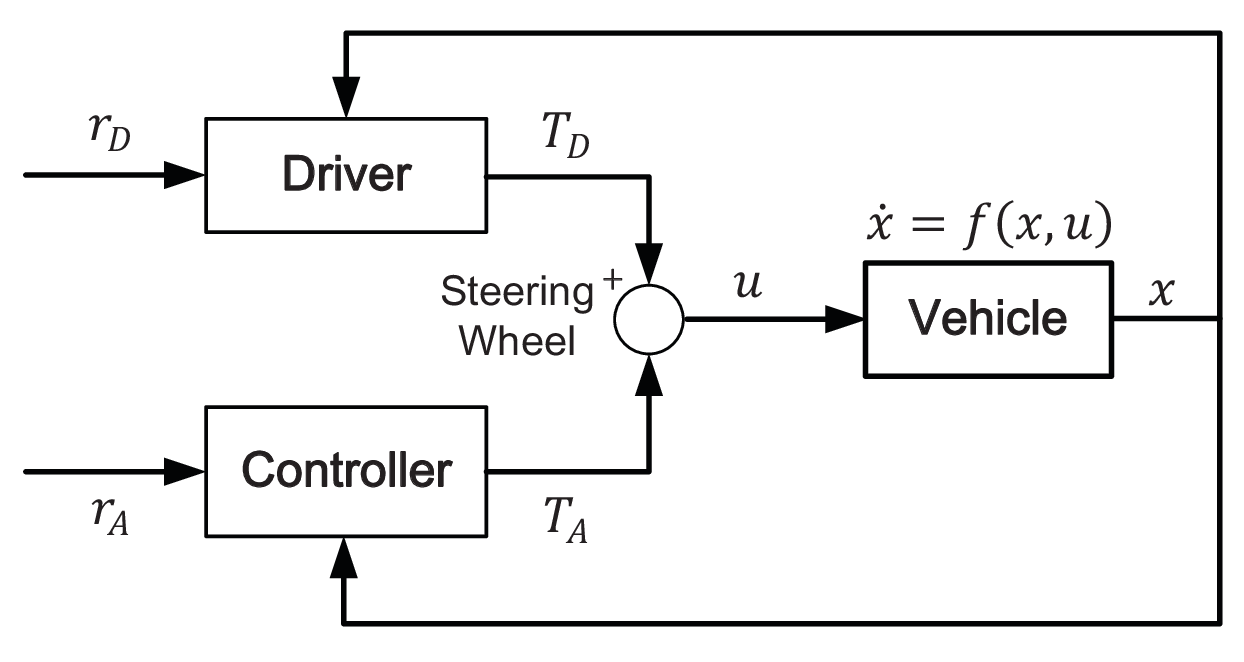}
      \label{fig.HSC}}
      \subfigure[]{
      \includegraphics[scale=0.6]{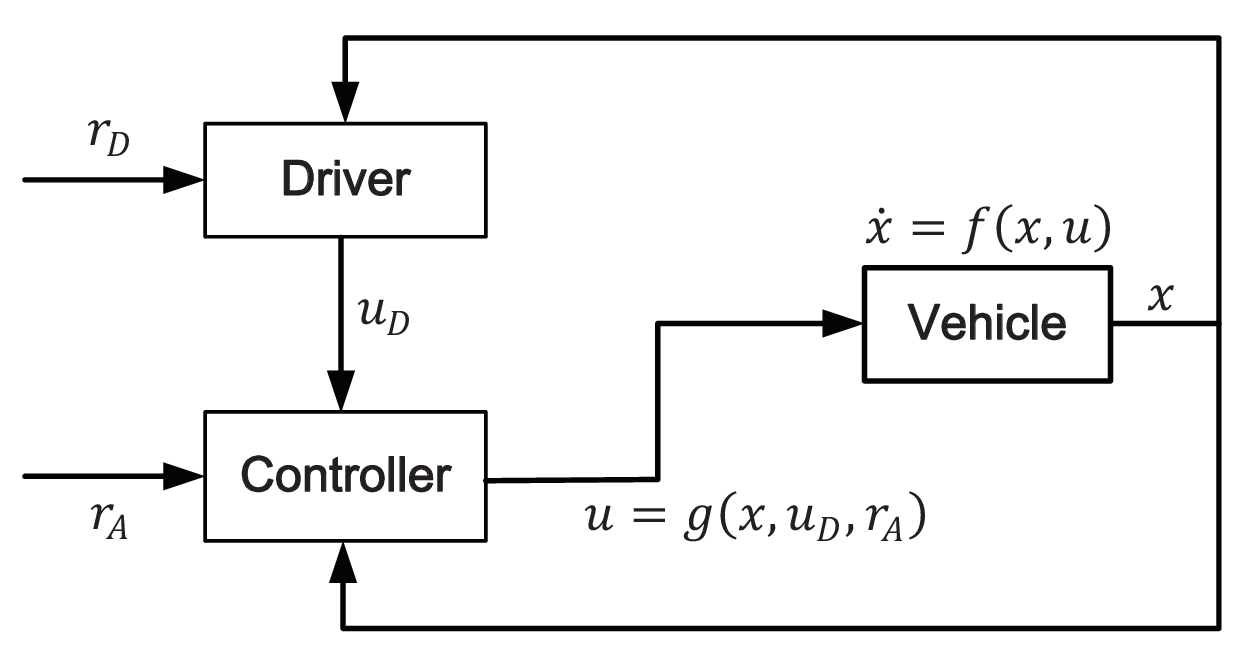}
      \label{fig.ISC}}
      \caption{Block diagrams of two shared control frameworks: a) haptic shared control; b) indirect shared control. }
      \label{fig.schemes}
\end{figure}

The frameworks of haptic shared control and indirect shared control are compared in Fig. \ref{fig.schemes}, where $r_D$ and $r_A$ are the reference paths of the driver and automation, respectively, $x$ is the vehicle state governed by the vehicle dynamics $\dot{x} = f(x, u)$, which is to be specified in Section \ref{sec.vehicle_dynamics}. In haptic shared control, the driver and controller simultaneously apply a control torque (denoted by $T_D$ and $T_A$) on the steering wheel and the final steering input $u$ (steering wheel angle) is determined by the resultant torque. The two parties jointly control the vehicle through physical interaction on the steering wheel. In indirect shared control, the steering wheel angle $u_D$ is solely commanded by the driver, whereas it can be transformed by the controller into the final steering input $u$ before being delivered to the vehicle. $u_D$ and $u$ are connected by the controller's transformation function $u = g(x, u_D, r_A)$. In other words, the driver input $u_D$ can be viewed as the intermediate input of the vehicle. The automation blends its own control objective with the driver input through the transformation $u = g(x, u_D, r_A)$. If the transformation function is $u \equiv u_D$, indirect shared control degrades to manual driving; if $u$ is independent of $u_D$, the vehicle becomes autonomous.

The intermediate input transformation $u = g(x, u_D, r_A)$ is the crucial part of indirect shared control, which exploits the automation's path-tracking ability and respects the driver input at the same time. To put it another way, the driver can indirectly influence the vehicle control through $u = g(x, u_D, r_A)$. The driver's control authority and responsibility depend on how largely $u$ relies on $u_D$. In this sense, the driver and the automation share the control authority and guide the vehicle in a cooperative fashion. This scheme guarantees that the driver is actively involved in the loop because he is obligated to convey control input throughout driving. Meanwhile, the driver's control effort can be partly relieved by virtue of the assistant controller.

\subsection{Vehicle Dynamics} \label{sec.vehicle_dynamics}
\begin{figure}[thpb]
    \centering
    \includegraphics[scale = 0.7]{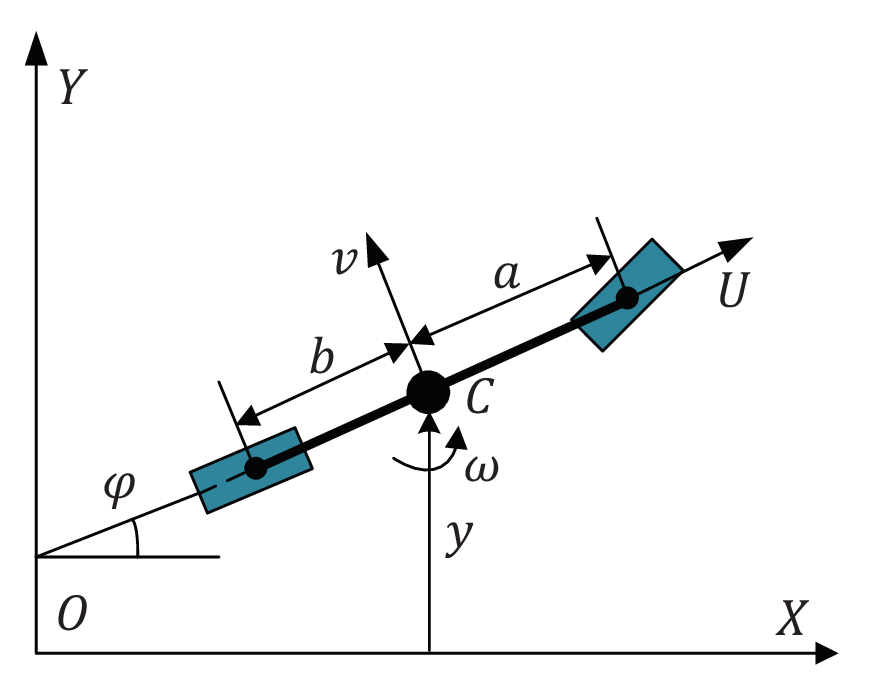}
    \caption{Linearized dynamic bicycle model.}
    \label{fig.vehiclemodel}
\end{figure}
The vehicle model used in this paper is the single-track dynamic bicycle model \cite{Cole06VSD} shown in Fig. \ref{fig.vehiclemodel}. This model has high fidelity if the vehicle sideslip angle is small, i.e., the tires work in a linear region. If the vehicle is running at a constant longitudinal velocity $U$ and the vehicle yaw angle from \emph{X}-axis is small, its dynamics can be described by the linear time-invariant equation
\be
\label{eq.vehicle}
\begin{split}
   \dot{x}(t) &= A_cx(t) + B_cu(t) \\
     z(t) &= C_cx(t)
\end{split}
\ee
where the state $x(t) = \begin{bmatrix} v(t) & \omega(t) & y(t) & \psi(t) \end{bmatrix}^\mathrm{T}$ ($v$: lateral velocity, $\omega$: yaw rate, $y$: lateral displacement, $\psi$: yaw angle), $u(t)=\delta(t)$ is the steering wheel angle, $z(t) = \begin{bmatrix} y(t) & \psi(t) \end{bmatrix}^\mathrm{T}$ is the output used to evaluate the path-following performance and provide feedback. The subscript ``c'' denotes ``continuous-time''. $A_c$ and $B_c$ are constant matrices related to the vehicle's intrinsic properties and the constant velocity $U$:
\be
\label{eq.dyna_matrices}
\begin{split}
A_c &=
\begin{bmatrix}[2]
  \dfrac{-(C_f+C_r)}{mU} & \dfrac{-(aC_f-bC_r)}{mU}-U & 0 & 0 \\
  \dfrac{-(aC_f-bC_r)}{I_zU} & \dfrac{-(a^2C_f+b^2C_r)}{I_zU} & 0 & 0 \\
  1 & 0 & 0 & U \\
  0 & 1 & 0 & 0
\end{bmatrix} \\
B_c &=
\begin{bmatrix}
  \dfrac{C_f}{i_sm} & \dfrac{aC_f}{i_sI_z} & 0 & 0
\end{bmatrix}^\mathrm{T}, \,\,
C_c = \begin{bmatrix}[1.5]
      0 & 0 & 1 & 0 \\
      0 & 0 & 0 & 1
    \end{bmatrix}
\end{split}
\ee
where $C_f$ is front cornering stiffness, $C_r$ rear cornering stiffness, $m$ the vehicle mass, $a$ the distance from center of mass to front axle, $b$ the distance from center of mass to rear axle, $I_z$ the polar moment of inertia, and $i_s$ the steering ratio. The continuous-time vehicle dynamics (\ref{eq.vehicle}) can be converted to the discrete-time equation
\be
\label{eq.vehicled}
\begin{split}
   x(k+1) &= A\,x(k) + Bu(k) \\
     z(k) &= Cx(k)
\end{split}
\ee
where $k$ denotes discrete time index.

\section{Indirect Shared Control}
\label{sec.strategies}
This section formulates the strategies of the controller and driver in indirect shared control. The controller's transformation strategy $u = g(x, u_D, r_A)$ is first introduced, which should ensure that the interests of both the driver and automation are respected. Second, given this transformation strategy, the driver's adaptive control strategy is modeled.

\subsection{Controller Strategy}
We assume that the controller can observe the driver input $u_D$ at every time. The controller needs to follow a reference path $r_A$ while considering the driver input. In practice, the reference path may be generated by a path planner, although this is beyond the scope of this paper. We follow the weighted summation method \cite{Manabu06IATSS} to blend the driver input with the automation's desired input:
\begin{equation}
\label{eq.linear_weight}
u(k) = \lambda_D \, u_D(k) + \lambda_A \, u_A(k), \, \quad \lambda_D, \lambda_A \geq 0 \,,
\end{equation}
where $u$ is the final input delivered to the vehicle, $u_D$ the instant driver input, $u_A$ the automation's desired input, and $\lambda_D, \lambda_A$ the authority weights assigned to the driver input and the automation's desired input, respectively. $\lambda_D + \lambda_A \equiv 1$ is often assumed to avoid conflict and force escalation, as well as to simplify the control authority management using only one parameter.

The second part of the controller strategy is to specify how to generate the desired input $u_A$ at each time. In this study, $u_A$ is solely based on the automation's reference path and control objective, which means that $u_A$ is independent of $u_D$. We adopt MPC to calculate $u_A$, which uses the vehicle dynamics (\ref{eq.vehicled}) as the predictor and repeatedly programs the optimal input sequence over a predictive horizon associated with a cost function:
\begin{subequations}
\label{eq.automation_strategy}
\begin{align}
u_A(k) &= \bm{\mathrm{u}_{A,(1)}}^*(k) \\
\bm{\mathrm{u}_A}^*(k) &= \mathop{\arg\min}_{\{u_A(k), u_A(k+1), \cdots, u_A(k+N_c-1)\}}J_A(k) \\
x(k+1) &= Ax(k) + Bu_A(k) \\
z(k) &= Cx(k)
\end{align}
\end{subequations}
where $u_A$ is the automation's desired input following MPC, $N_c$ the control horizon, $\bm{\mathrm{u}_A}^*$ the optimal input sequence over the control horizon, and $J_A$ the automation's cost function. $\bm{\mathrm{u}_{A,(1)}}^*$ denotes the first element of $\bm{\mathrm{u}_A}^*$ used as the command at each time. In this paper, we consider the following quadratic cost to minimize the tracking error and effort:
\begin{multline}
\label{eq.automation_cost_function}
  J_A(k) \equiv \sum_{i = 1}^{N_p} \left \lVert z(k+i)-r_A(k+i) \right \rVert _{Q_A}^2 \\+ \sum_{i = 0}^{N_u - 1} \left \lVert u_A(k+i) \right \rVert _{R_A}^2
\end{multline}
where $r_A(k)$ is the automation's reference path at each time step, $N_p$ the predictive horizon, and $Q_A, R_A$ constant positive definite weighting matrices of appropriate dimensions. 

\subsection{Driver Strategy}
The driver strategy is modeled considering two main assumptions:
\begin{enumerate}
  \item The driver has identified the controller's transformation strategy through motor learning.
  \item The driver incorporates learned transformation strategy into their internal model for predictive control.
\end{enumerate}
The above assumptions state that the driver employs MPC-type strategy to perform path-following task \cite{Cole06VSD} but will adapt the controller's transformation strategy (\ref{eq.linear_weight}) into the predictor after motor learning. Accordingly, the driver's control strategy is modeled as
\begin{subequations}\label{eq.driver_strategy}
\begin{align}
 u_D(k) &= \bm{\mathrm{u}_{D,(1)}}^*(k)\\
 \bm{\mathrm{u}_D}^*(k) &= \mathop{\arg\min}_{\{u_D(k),u_D(k+1),\cdots,u_D(k+N_c-1)\}}J_D(k) \\
 \label{eq.driver_predictor_1}
 \text{s.t.} \; u(k) &= \lambda_Du_D(k)+\lambda_A u_A(k) \\
 \label{eq.driver_predictor_2}
x(k+1) &= Ax(k) + Bu(k) \\
z(k) &= Cx(k)
\end{align}
\end{subequations}
where $u_D$ is the driver input following MPC strategy, $N_c$ the control horizon which is assumed identical to the controller, $\bm{\mathrm{u}_D}^*$ is the optimal input sequence up to the control horizon, and $J_D$ is the driver's cost function. The key difference between (\ref{eq.driver_strategy}) and the MPC strategy for manual driving in \cite{Cole06VSD} is that the driver hereby incorporates the controller's transformation strategy (\ref{eq.linear_weight}) into the prediction.

Cost function $J_D$ is critical for modeling the driver behavior in indirect shared control. According to \cite{Cole06VSD, Burdet13MITpress}, the cost function for manual path following takes a quadratic form similar to (\ref{eq.automation_cost_function}). In this paper, we assume that the driver's cost function preserves the form in indirect shared control, i.e.,
\begin{multline}
\label{eq.driver_cost_function}
J_D(k) \equiv \sum_{i = 1}^{N_p} \left \lVert z(k+i)-r_D(k+i) \right \rVert _{Q_D}^2  \\
+ \sum_{i = 0}^{N_u - 1} \left \lVert u_D(k+i) \right \rVert _{R_D}^2
\end{multline}
where $r_D(k)$ is the driver's reference path at each step, $N_p$ the predictive horizon which is assumed identical to the controller, and $Q_D$ and $R_D$ constant positive definite weighting matrices of appropriate dimensions. Similar to the automation, the driver also seeks to minimize the path-tracking error and control effort.

The authority weights $\lambda_D$ and $\lambda_A$ determine the control authorities of the driver and automation. If one of them is set to zero, indirect shared control becomes either manual driving or fully autonomous driving:
\begin{enumerate}
  \item $\lambda_D = 0, \lambda_A = 1$ means that the controller does not take the driver input into account, i.e., the driver has no control authority, and the vehicle becomes autonomous.
  \item $\lambda_D = 1, \lambda_A = 0$ means that the controller completely complies with the driver, i.e., the driver retains direct control authority, and the vehicle becomes manually driven.
\end{enumerate}

\section{Analytical Solution} \label{sec.derivation}
In this section, we derive the analytic expressions of the strategies, i.e., $u_A(k)$, $u_D(k)$, and $u(k)$, formulated in Section \ref{sec.strategies}. By doing this, we can better observe how indirect shared control shapes the driver's control strategy in comparison with manual driving cases. More specifically, we can obtain the influence of authority weights $\lambda_D$ and $\lambda_A$ on the driver behavior, which can be used to improve the design of indirect shared control. The calculation of the automation's desired input $u_A(k)$ is actually an unconstrained MPC-tracking problem. We directly give its analytic expression based on existing theoretical solution. The focus of this section lies in the derivation of the driver strategy $u_D(k)$ from which $u_A(k)$ and thus $u(k)$ can be computed through (\ref{eq.linear_weight}). For real-time implementation, there are some efficient MPC computation methods such as in \cite{Li15TITS,Li16Neuro}.

For simplicity, we set $N_p \equiv N_c \equiv N$, and introduce several notations to make the derivation more concise:
\ben
\bm{\mathrm{x}}(k) =
\begin{bmatrix}
  x(k+1) \\
  x(k+2) \\
  \vdots \\
  x(k+N)
\end{bmatrix}, \qquad
\bm{\mathrm{u}_A}(k) =
\begin{bmatrix}
  u_A(k) \\
  u_A(k+1) \\
  \vdots \\
  u_A(k+N-1)
\end{bmatrix}
\een
\ben
\bm{\mathrm{u}_D}(k) =
\begin{bmatrix}
  u_D(k) \\
  u_D(k+1) \\
  \vdots \\
  u_D(k+N-1)
\end{bmatrix}, \quad
\bm{\mathrm{z}}(k)=
\begin{bmatrix}
  z(k+1) \\
  z(k+2) \\
  \vdots \\
  z(k+N)
\end{bmatrix}
\een
\ben
\bm{\mathrm{r}_A}(k) =
\begin{bmatrix}
  r_A(k+1) \\
  r_A(k+2) \\
  \vdots \\
  r_A(k+N)
\end{bmatrix}, \qquad
\bm{\mathrm{r}_D}(k)=
\begin{bmatrix}
  r_D(k+1) \\
  r_D(k+2) \\
  \vdots \\
  r_D(k+N)
\end{bmatrix}
\een
\ben
\mathcal{A} =
\begin{bmatrix}
  A \\
  A^2 \\
  \vdots \\
  A^N
\end{bmatrix}, \qquad
\mathcal{B} =
\begin{bmatrix}
  B & 0 & \cdots & 0 \\
  AB & B & \cdots & 0 \\
  \vdots & \vdots & \ddots & \vdots \\
  A^{N-1}B & A^{N-2}B & \cdots & B
\end{bmatrix}
\een
\ben
\mathcal{C} =
\begin{bmatrix}
C &  &  &  \\
& C &  &  \\
&  & \ddots &  \\
&  &  & C
\end{bmatrix} \!
\left. \vphantom{
\begin{bmatrix}
0\\0\\0\\0
\end{bmatrix}}\right\}\! N, \quad
\Phi = \mathcal{C}\mathcal{A} =
\begin{bmatrix}
CA \\
CA^2  \\
\vdots  \\
CA^{N-1}
\end{bmatrix}
\een
\ben
\Theta = \mathcal{C}\mathcal{B} =
\begin{bmatrix}
  CB & 0 & \cdots & 0 \\
  CAB & CB & \cdots & 0 \\
  \vdots & \vdots & \ddots & \vdots \\
  CA^{N-1}B & CA^{N-2}B & \cdots & CB
\end{bmatrix}
\een
\ben
\mathcal{Q}_A\,(\mathcal{Q}_D) =
\begin{bmatrix}
Q_A\,(Q_D) &  &  &  \\
& Q_A\,(Q_D) &  &  \\
&  & \ddots &  \\
&  &  & Q_A\,(Q_D)
\end{bmatrix} \!
\left. \vphantom{
\begin{bmatrix}
0\\0\\0\\0
\end{bmatrix}}\right\}\! N
\een
\ben
\mathcal{R}_A \,(\mathcal{R}_D)=
\begin{bmatrix}
R_A\,(R_D) &  &  &  \\
& R_A\,(R_D) &  &  \\
&  & \ddots &  \\
&  &  & R_A\,(R_D)
\end{bmatrix}\!
\left. \vphantom{
\begin{bmatrix}
0\\0\\0\\0
\end{bmatrix}}\right\}\! N
\een
We first present the analytical solution of $u_A(k)$ to the MPC problem (\ref{eq.automation_strategy},\ref{eq.automation_cost_function}). This is an unconstrained MPC-tracking problem with solution
\be
\label{eq.u_A(k)}
u_A(k) = e_1^{\mathrm{T}}\mathcal{K}_A\bm{\varepsilon_A}(k) \, .
\ee
where $e_1^{\mathrm{T}} \equiv \overbrace{\begin{bmatrix} 1 & 0 & \cdots & 0 \end{bmatrix}}^N$,
$ \mathcal{K}_A \equiv
\begin{bmatrix}
  \sqrt{\mathcal{Q}_A}\Theta \\
  \sqrt{\mathcal{R}_A}
\end{bmatrix}^{\dagger}
\begin{bmatrix}
  \sqrt{\mathcal{Q}_A}\\
  0
\end{bmatrix} $
(`$\dagger$' denotes pseudo-inverse) is a constant matrix and $\bm{\varepsilon_A}(k) \equiv \bm{\mathrm{r}_A}(k)-\Phi x(k)$ is associated with the reference path up to the predictive horizon $\bm{\mathrm{r}_A}(k)$ and the current vehicle state $x(k)$.

The focus of this section is to derive $u_D(k)$, given the problem (\ref{eq.driver_strategy}, \ref{eq.driver_cost_function}) and the solution of $u_A(k)$ in (\ref{eq.u_A(k)}). Substituting (\ref{eq.u_A(k)}) into (\ref{eq.driver_predictor_1}), and then into (\ref{eq.driver_predictor_2}), we can rewrite the driver's prediction equation as
\begin{multline}
x(k+1) = Ax(k) + \lambda_DBu_D(k) \\
+\lambda_ABe_1^{\mathrm{T}}\mathcal{K}_A[\bm{\mathrm{r_A}}(k)-\Phi x(k)].
\end{multline}
Rearranging the terms, we have
\begin{multline}
\label{eq.new_prediction_1}
x(k+1) = (A-\lambda_ABe_1^{\mathrm{T}}\mathcal{K}_A\Phi)x(k) + \lambda_D B u_D(k) \\
+\lambda_ABe_1^{\mathrm{T}}\mathcal{K}_A\bm{\mathrm{r_A}}(k).
\end{multline}
Denoting
\be
\tilde{A} \equiv A-\lambda_ABe_1^{\mathrm{T}}\mathcal{K}_A\Phi
\ee
and
\be
w_A(k) \equiv e_1^{\mathrm{T}}\mathcal{K}_A\bm{\mathrm{r}_A}(k) \, ,
\ee
we can rewrite (\ref{eq.new_prediction_1}) as
\be
\label{eq.new_prediction_2}
x(k+1) = \tilde{A}x(k)+\lambda_DBu_D(k) +\lambda_ABw_A(k).
\ee
Iterating (\ref{eq.new_prediction_2}) for $N-1$ times and stacking the results, yield
\be
\label{eq.expanded_large}
\begin{split}
&\begin{bmatrix}
  x(k+1) \\
  x(k+2) \\
  \vdots \\
  x(k+N)
\end{bmatrix} =
\begin{bmatrix}
  \tilde{A} \\
  \tilde{A}^2 \\
  \vdots \\
  \tilde{A}^N
\end{bmatrix} x(k) \\
&+\lambda_D
\begin{bmatrix}
  B & 0 & \cdots & 0 \\
  \tilde{A}B & B & \cdots & 0 \\
  \vdots & \vdots & \ddots & \vdots \\
  \tilde{A}^{N-1}B & \tilde{A}^{N-2}B & \cdots & B
\end{bmatrix}
\begin{bmatrix}
  u_D(k) \\
  u_D(k+1) \\
  \vdots \\
 u_D(k+N-1)
\end{bmatrix} \\
&+\lambda_A
\begin{bmatrix}
  B & 0 & \cdots & 0 \\
  \tilde{A}B & B & \cdots & 0 \\
  \vdots & \vdots & \ddots & \vdots \\
  \tilde{A}^{N-1}B & \tilde{A}^{N-2}B & \cdots & B
\end{bmatrix}
\begin{bmatrix}
  w_A(k) \\
  w_A(k+1) \\
  \vdots \\
 w_A(k+N-1)
\end{bmatrix}
\end{split}
\ee
Using the notations
\ben
\bm{\mathrm{w}_A}(k) =
\begin{bmatrix}
  w_A(k+1) \\
  w_A(k+2) \\
  \vdots \\
  w_A(k+N)
\end{bmatrix}, \qquad
\tilde{\mathcal{A}} =
\begin{bmatrix}
  \tilde{A} \\
  \tilde{A}^2 \\
  \vdots \\
  \tilde{A}^N
\end{bmatrix}
\een
\ben
\tilde{\mathcal{B}} =
\begin{bmatrix}
  B & 0 & \cdots & 0 \\
  \tilde{A}B & B & \cdots & 0 \\
  \vdots & \vdots & \ddots & \vdots \\
  \tilde{A}^{N-1}B & \tilde{A}^{N-2}B & \cdots & B
\end{bmatrix},
\tilde{\Phi} = \mathcal{C}\tilde{\mathcal{A}} =
\begin{bmatrix}
  C\tilde{A} \\
  C\tilde{A}^2 \\
  \vdots \\
  C\tilde{A}^N
\end{bmatrix}
\een
\ben
\tilde{\Theta} = \mathcal{C}\tilde{\mathcal{B}} =
\begin{bmatrix}
  CB & 0 & \cdots & 0 \\
  C\tilde{A}B & CB & \cdots & 0 \\
  \vdots & \vdots & \ddots & \vdots \\
  C\tilde{A}^{N-1}B & C\tilde{A}^{N-2}B & \cdots & CB
\end{bmatrix}.
\een
Eq. (\ref{eq.expanded_large}) can be rewritten as
\be
\bm{\mathrm{x}}(k) = \tilde{\mathcal{A}}x(k)+\lambda_D\tilde{\mathcal{B}}\bm{\mathrm{u}_D}(k)
+\lambda_A\tilde{\mathcal{B}}\bm{\mathrm{w}_A}(k)
\ee
and the output is
\be
\begin{split}
\label{eq.z(k)}
\bm{\mathrm{z}}(k) &= \mathcal{C}\bm{\mathrm{x}}(k) = \mathcal{C}\tilde{\mathcal{A}}x(k)+\lambda_D\mathcal{C}\tilde{\mathcal{B}}\bm{\mathrm{u}_D}(k)
+\lambda_A\mathcal{C}\tilde{\mathcal{B}}\bm{\mathrm{w}_A}(k) \\
&= \tilde{\Phi}x(k)+\lambda_D\tilde{\Theta}\bm{\mathrm{u}_D}(k)+\lambda_A\tilde{\Theta}\bm{\mathrm{w}_A}(k).
\end{split}
\ee
The driver's cost function (\ref{eq.driver_cost_function}) can be expressed as
\be
J_D(k) = \left \lVert \bm{\mathrm{z}}(k)-\bm{\mathrm{r}_D}(k) \right \rVert _{\mathcal{Q}_D}^2
+\left \lVert \bm{\mathrm{u}_D}(k) \right \rVert _{\mathcal{R}_D}^2.
\ee
According to (\ref{eq.z(k)}), the path-tracking error can be stated as
\be
\bm{\mathrm{z}}(k)-\bm{\mathrm{r}_D}(k) =\tilde{\Phi}x(k)+\lambda_D\tilde{\Theta}\bm{\mathrm{u}_D}(k)+\lambda_A\tilde{\Theta}\bm{\mathrm{w}_A}(k)-
\bm{\mathrm{r}_D}(k).
\ee
Let us set
\be
\bm{\varepsilon_D}(k) \equiv \bm{\mathrm{r}_D}(k)-\tilde{\Phi}x(k)-\lambda_A\tilde{\Theta}\bm{\mathrm{w}_A}(k) \, ,
\ee
then we have
\be
\bm{\mathrm{z}}(k)-\bm{\mathrm{r}_D}(k) = \lambda_D\tilde{\Theta}\bm{\mathrm{u}_D}(k)-\bm{\varepsilon_D}(k) \, .
\ee
The driver's cost function can be then expressed as
\be
\begin{split}
\label{eq.driver_cost_function_2}
J_D(k) &= \left \lVert \lambda_D\tilde{\Theta}\bm{\mathrm{u}_D}(k)-\bm{\varepsilon_D}(k) \right \rVert _{\mathcal{Q}_D}^2
+ \left \lVert \bm{\mathrm{u}_D}(k) \right \rVert _{\mathcal{R}_D}^2 \\
&=
\begin{Vmatrix}
  \sqrt{\mathcal{Q}_D}[\lambda_D\tilde{\Theta}\bm{\mathrm{u}_D}(k)-\bm{\varepsilon_D}(k)] \\
  \sqrt{\mathcal{R}_D}\bm{\mathrm{u}_D}(k)
\end{Vmatrix}^2.
\end{split}
\ee
The receding optimization problem (\ref{eq.driver_strategy}) is converted to
\be
\label{eq.receiding_optimization}
\bm{\mathrm{u}_D}^*(k) = \mathop{\arg\min}_{\bm{\mathrm{u}_D}(k)}
\begin{Vmatrix}
  \sqrt{\mathcal{Q}_D}[\lambda_D\tilde{\Theta}\bm{\mathrm{u}_D}(k)-\bm{\varepsilon_D}(k)] \\
  \sqrt{\mathcal{R}_D}\bm{\mathrm{u}_D}(k)
\end{Vmatrix}^2.
\ee
The solution of (\ref{eq.receiding_optimization}) is identical to the least-square solution of the problem
\be \label{eq.least_square}
\begin{bmatrix}
  \lambda_D\sqrt{\mathcal{Q}_D}\tilde{\Theta} \\
  \sqrt{\mathcal{R}_D}
\end{bmatrix}
\bm{\mathrm{u}_D}(k) =
\begin{bmatrix}
  \sqrt{\mathcal{Q}_D}\\
  0
\end{bmatrix}
\bm{\varepsilon_D}(k),
\ee
which can be stated as
\be
\bm{\mathrm{u}_D}^*(k) =
\begin{bmatrix}
  \lambda_D\sqrt{\mathcal{Q}_D}\tilde{\Theta} \\
  \sqrt{\mathcal{R}_D}
\end{bmatrix}^{\dagger}
\begin{bmatrix}
  \sqrt{\mathcal{Q}_D}\\
  0
\end{bmatrix} \bm{\varepsilon_D}(k)
=\mathcal{K}_D\bm{\varepsilon_D}(k),
\ee
where $\mathcal{K}_D \equiv
\begin{bmatrix}
  \lambda_D\sqrt{\mathcal{Q}_D}\tilde{\Theta} \\
  \sqrt{\mathcal{R}_D}
\end{bmatrix} ^{\dagger}
\begin{bmatrix}
  \sqrt{\mathcal{Q}_D}\\
  0
\end{bmatrix}$
is a constant matrix, $\bm{\varepsilon_D}(k)$ is a vector associated with the driver's reference path up to the predictive horizon $\bm{\mathrm{r}_D}(k)$, the current vehicle state $x(k)$, and $\bm{\mathrm{w}_A}(k)$. Note that
\be
\bm{\mathrm{w}_A}(k) =
\begin{bmatrix}
  w_A(k) \\
  w_A(k+1) \\
  \vdots \\
  w_A(k+N-1)
\end{bmatrix} =
\begin{bmatrix}
  e_1^{\mathrm{T}}\mathcal{K}_A\bm{\mathrm{r_A}}(k) \\
  e_1^{\mathrm{T}}\mathcal{K}_A\bm{\mathrm{r_A}}(k+1) \\
  \vdots \\
  e_1^{\mathrm{T}}\mathcal{K}_A\bm{\mathrm{r_A}}(k+N-1)
\end{bmatrix},
\ee
which actually contains the automation's reference path up to horizon $2N-1$, because $\bm{\mathrm{r_A}}(k+N-1)$ encompasses $r_A(k+N-1),\cdots,r_A(k+2N-1)$.

Finally, the driver's control strategy is to use the first element of $\bm{\mathrm{u}_D}^*(k)$ as motor command, which is
\be
\label{eq.driver_strategy_expression}
u_D(k) = e_1^{\mathrm{T}}\mathcal{K}_D\bm{\varepsilon_D}(k) \, .
\ee
We can then compute $u(k)$ using equations (\ref{eq.linear_weight}, \ref{eq.u_A(k)}, \ref{eq.driver_strategy_expression}):
\begin{equation}
u(k) = \lambda_De_1^{\mathrm{T}}\mathcal{K}_D\bm{\varepsilon_D}(k)+\lambda_Ae_1^{\mathrm{T}}\mathcal{K}_A\bm{\varepsilon_A}(k).
\end{equation}

We can briefly validate (\ref{eq.driver_strategy_expression}) by examining two special cases, i.e., manual driving ($\lambda_D = 1$, $\lambda_A = 0$) and fully automated driving ($\lambda_D = 0$, $\lambda_A = 1$):
\begin{enumerate}
  \item
  Let $\lambda_D = 1$ and $\lambda_A = 0$, which transforms indirect shared control into manual driving.
In this case, $\tilde{A}=A$, then $\tilde{\mathcal{A}}=\mathcal{A}$, $\tilde{\mathcal{B}}=\mathcal{B}$, $\tilde{\Phi}=\Phi$, $\tilde{\Theta}=\Theta$.
$\mathcal{K}_D =
\begin{bmatrix}
  \sqrt{\mathcal{Q}_D}\Theta \\
  \sqrt{\mathcal{R}_D}
\end{bmatrix} ^{\dagger}
\begin{bmatrix}
  \sqrt{\mathcal{Q}_D}\\
  0
\end{bmatrix}$, $\bm{\varepsilon_D}(k) = \bm{\mathrm{r}_D}(k)-\Phi x(k)$. The strategy (\ref{eq.driver_strategy_expression}) becomes the manual driving strategy in \cite{Cole06VSD}. This is consistent with our intuition.
  \item
  Let $\lambda_D = 0$ and $\lambda_A = 1$, which transforms indirect shared control into automated driving. In this case, according to (\ref{eq.driver_cost_function_2}), the driver's cost function becomes $J_D(k) =
   \left \lVert -\bm{\varepsilon_D}(k) \right \rVert _{\mathcal{Q}_D}^2
+ \left \lVert \bm{\mathrm{u}_D}(k) \right \rVert _{\mathcal{R}_D}^2$.
Because $\bm{\varepsilon_D}(k)$ is not subject to $\bm{\mathrm{u}_D}(k)$, the driver can only minimize $\left \lVert \bm{\mathrm{u}_D}(k) \right \rVert _{\mathcal{R}_D}^2$, which yields the trivial solution $\bm{\mathrm{u}_D}^*(k) = 0$ and $u_D(k) = 0$. This means that if the driver realizes that he has no control authority, he will not take any action but stays relaxed, which is consistent with our intuition.
\end{enumerate}

\section{Automatic Weight Switching} \label{sec.weight_switching}
Intuitively, the authority weights $\lambda_D$ and $\lambda_A$ should vary with the specific situation. When the automation is consistent with the driver intention, i.e., they are following an identical path, $\lambda_A$ could be smaller ($\lambda_D$ could be larger) to relieve the driver. However, if the driver intends to divert the vehicle from the original path, $\lambda_D$ should increase ($\lambda_A$ should be decreased) in order to follow the driver intention. A typical situation is when there is an obstacle ahead of the vehicle that is undetected by the automation, the driver should be assigned a larger $\lambda_D$ (i.e., a smaller $\lambda_A$) in order to avoid the obstacle easily.

A simple approach to achieve this objective would be to allow the driver to set the authority weights manually, for example by pressing a button to inform the automation that more control authority is required. However there may not be sufficient time for the driver to do the notification, and this manual setting approach places a heavy burden to the driver if the request is frequent. Therefore, it is desirable that the authority weights $\lambda_D$ and $\lambda_A$ can change automatically according to the driver intention.

We hereby use a model-based method to detect whether the automation corresponds to the driver intention. We assume that the driver's reference path is identical to the automation, i.e., $\bm{r_D}\equiv\bm{r_A}$, if their intentions are matched, and the automation knows the driver's $Q_D$ through either prior knowledge or real-time estimation. Therefore, the automation can estimate the expected driver input at every time step based on the driver model (\ref{eq.driver_strategy_expression}):
\begin{equation} \label{eq.expected_driver_input}
\hat u_D(k) =  e_1^{\mathrm{T}}\mathcal{K}_D\bm{\varepsilon_D}(k).
\end{equation}
In reality, when the driver intention is consistent with the automation, this should result in some oscillatory error between the actual driver input and expected one, known as the model error. However, if the driver intention becomes inconsistent with the automation, this error steadily increases in one direction. To detect such inconsistency between the driver and automation we thus monitor the mean cumulative error within a constant-length sliding window:
\begin{equation} \label{eq.error_definition}
\delta(k) = \frac{1}{H}\left|\sum_{j=k-H+1}^k \Big(\tilde u_D(j)-\hat u_D(j)\Big)\right|,
\end{equation}
where $H$ is the window length, and $\tilde u_D$ the actual driver input. A threshold-based rule is adopted to detect the driver intention and switch the weights accordingly:
\begin{equation} \label{eq.incons_detection}
\begin{cases}
\lambda_D(k+1) =\lambda_D^+ \,,\; \lambda_A(k+1) =\lambda_A^-& \text{if $\delta(k) \geq \delta^*$} \\
\lambda_D(k+1) =\lambda_D^-\,, \;\lambda_A(k+1) =\lambda_A^+& \text{if $\delta(k) < \delta^*$}
\end{cases},
\end{equation}
where $\lambda_D^+$ ($\lambda_A^+$) and $\lambda_D^-$ ($\lambda_A^-$) denote the constant larger and smaller weights of the driver (automation), respectively. $\delta^*$ is a predefined threshold to judge if the driver intention is still consistent with the automation. This rule indicates that if the mean cumulative error between $\tilde u_D(k)$ and $\hat u_D(k)$ within a constant-length window exceeds a certain threshold, the driver is deemed to hold a different intention with the automation, and the authority weights will be switched accordingly.

The selection of $\delta^*$ should rely on the prior knowledge of the model error and inevitably raises some trade-off. If $\delta^*$ is too large, the detection becomes increasingly insensitive such that there is a latency of weight switching in response to the driver's intention change; otherwise if $\delta^*$ is too small, the detection is unnecessarily sensitive which leads to frequent false switching. In practice, $\delta^*$ should be tuned until satisfactory performance is achieved; it will typically depend on each individual.

\section{Simulation} \label{sec.simulation}
To validate the proposed approach, we carried out simulations of indirect shared control in two scenarios: path following and obstacle avoidance. In the path-following scenario, the controller assists the driver to follow the reference path (usually designated as the lane centerline). It is expected that under such a circumstance indirect shared control can reduce the driver's control effort and enhance the vehicle path-tracking performance. In the obstacle-avoidance scenario, the driver attempts to control the vehicle in order to avoid an obstacle which is undetected by the automation. Indirect shared control enables the driver to avoid the obstacle without turning off the automation. In each scenario, the model proposed in this paper is compared with a conventional driver model without adaptation. Finally, this section compares how automatic weight switching facilitates the obstacle-avoidance task relative to static weighting.

The parameters for simulation are given in Table \ref{tab.parameters}, which include the parameters of vehicle dynamics and of the driver's (automation's) predictive control. Note that $Q_A$ is larger than $Q_D$ (PF) because the automation generally has superior path tracking capability to a human driver. Moreover, a small $Q_D$ (PF) also indicates that the driver tends to relax the control by compromising tracking performance in path-following tasks. Nevertheless, $Q_D$ (OA) is much larger than $Q_D$ (PF) because in emergency situations the driver prefers to guide the vehicle along his desired path, even at the cost of a large control effort.
\begin{table}
\centering
\caption{Simulation parameters}
\label{tab.parameters}
{
\setlength{\extrarowheight}{1.5pt}
\begin{tabular}{C{3.5cm}cc}
 \hline
 Front wheel cornering stiffness &$C_f$ & 12000 [N/rad]  \\
 Rear wheel cornering stiffness &$C_r$ & 8000 [N/rad] \\
 Distance from mass center to front axle &$a$ & 0.92 [m] \\
 Distance from mass center to rear axle &$b$ & 1.38 [m] \\
 Mass &$m$ & 1200 [kg] \\
 Polar moment of inertia &$I_z$ & 1500 [kg$\cdot\mathrm{m}^2$] \\
 Steering ratio &$i_s$ & 16 \\
 Longitudinal velocity &$U$ & 20 [m/s] \\
 Sampling time &$T$ & 0.02 [s] \\
 Predictive horizon &$N$ & 50 \\
 Automation weighting matrix &$Q_A$ &
 $\begin{bmatrix}
    1.5 &  \\
     & 0.6
\end{bmatrix}$ \\
Driver weighting matrix in path following &$Q_D$ (PF) &
 $\begin{bmatrix}
    0.036 &  \\
     & 0.02
\end{bmatrix}$ \\
Driver weighting matrix in obstacle avoidance &$Q_D$ (OA) &
 $\begin{bmatrix}
    36 &  \\
     & 20
\end{bmatrix}$ \\
\hline
\end{tabular}
}
\end{table}

\subsection{Path Following} \label{sec.PF}
\begin{figure}[thpb]
      \centering
      \subfigure[]{
      \includegraphics[scale=0.5]{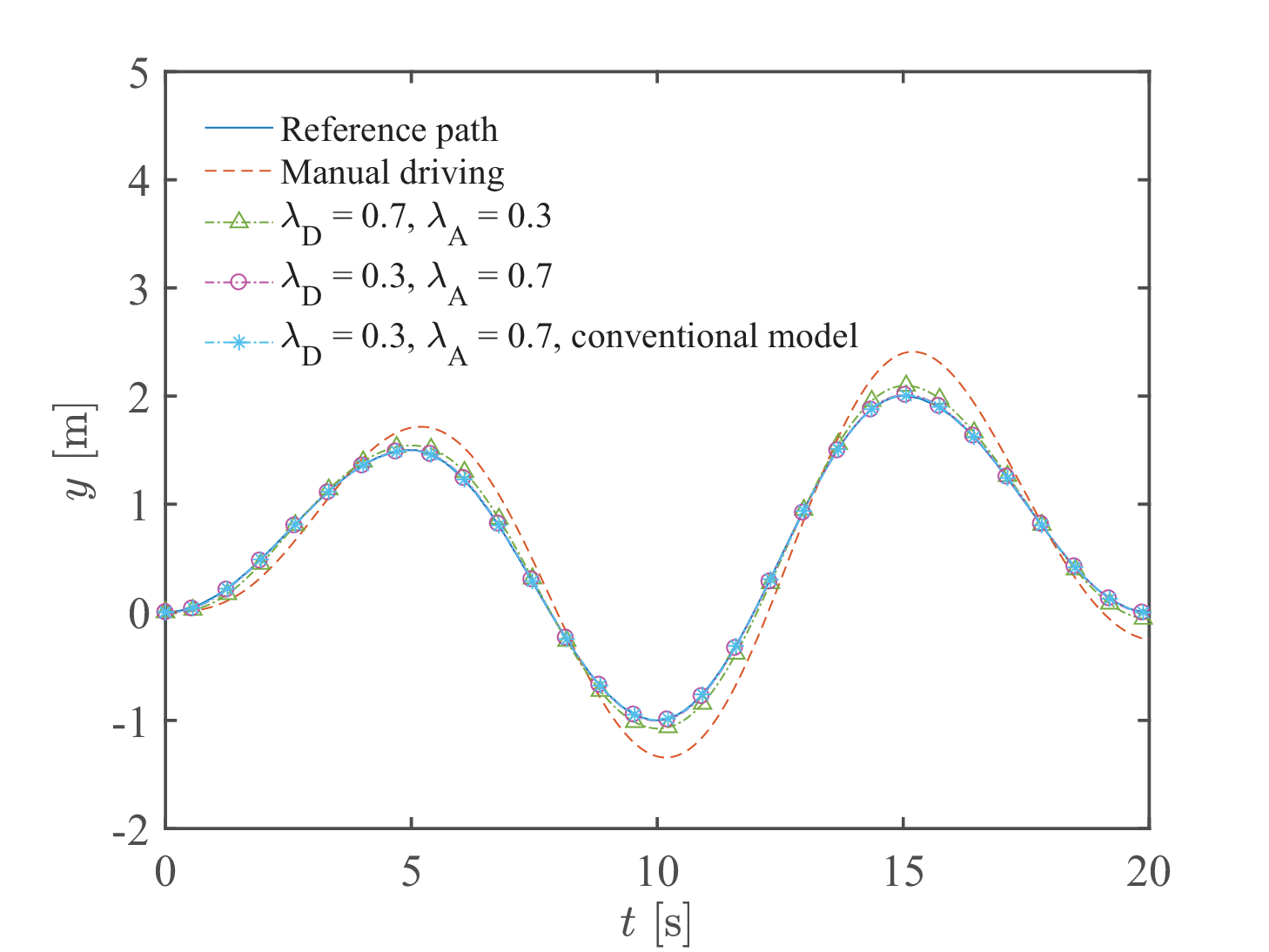}
      \label{fig.PF_y}}
      \subfigure[]{
      \includegraphics[scale=0.5]{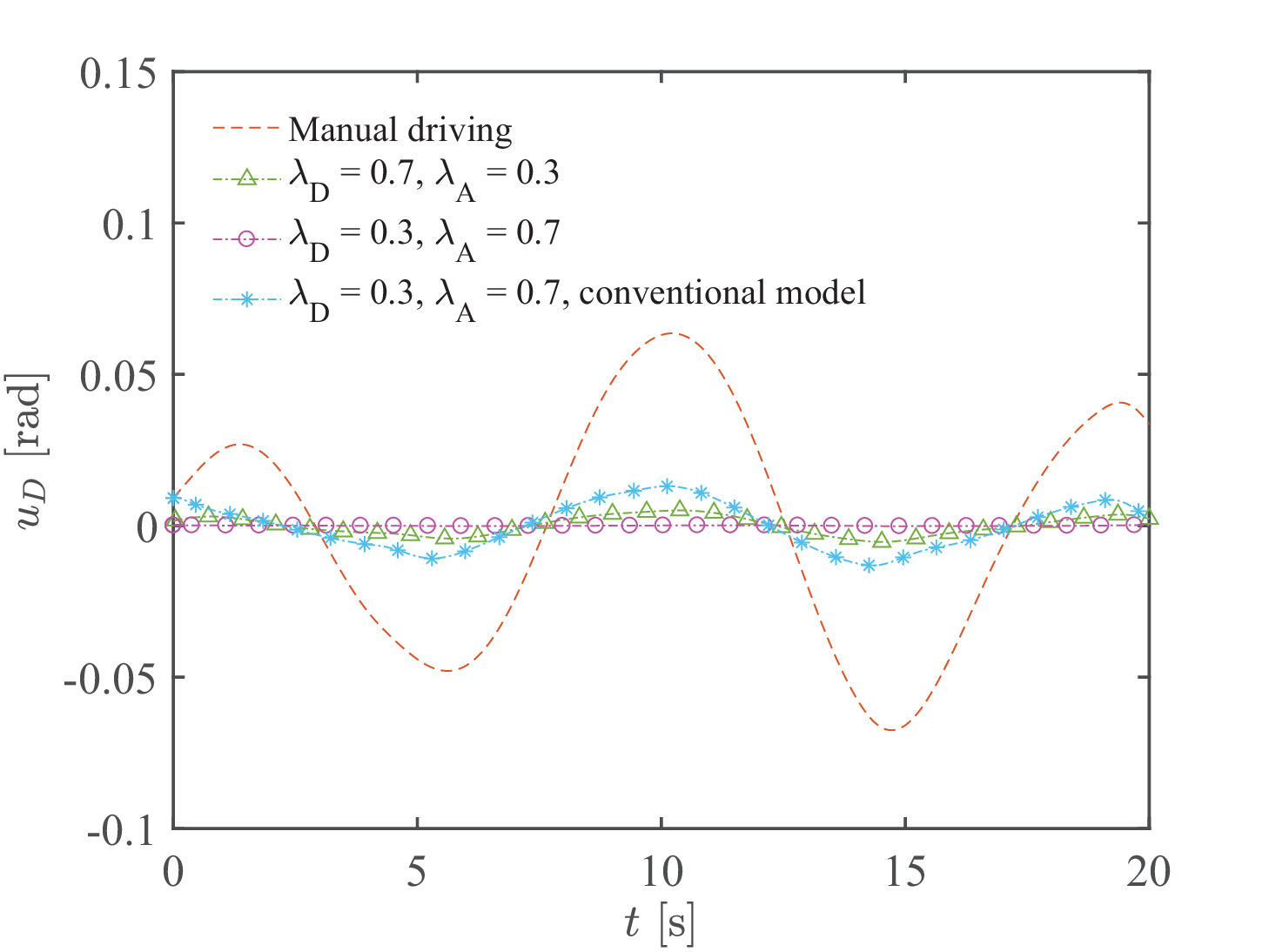}
      \label{fig.PF_u_D}}
      \caption{Path-following scenario: a) vehicle lateral displacement; b) driver's control effort. }
      \label{fig.PF}
\end{figure}

This part examines the benefits of indirect shared control in a path-following scenario. Fig. \ref{fig.PF_y} illustrates the trajectory of lateral displacement $y$ in different degrees of shared control (manual driving is a special case of shared control with $\lambda_D = 1$ and $\lambda_A = 0$). Corresponding driver control effort is depicted in Fig. \ref{fig.PF_u_D}. The reference path is designed as a continuous curve. According to our assumption, the driver tracks the same reference path as the automation. Additionally, the driver model without adaptation (denoted as ``conventional model'' in the figures) is also investigated. This is done by setting $\lambda_D = 1$ and $\lambda_A = 0$ when calculating the driver input according to (\ref{eq.driver_strategy_expression}). It can be seen in Fig. \ref{fig.PF_y} that the vehicle path-tracking performance is enhanced as the automation's control authority $\lambda_A$ increases. The model difference is not evident in terms of the path-tracking error. Fig. \ref{fig.PF_u_D} shows that the driver's control effort is significantly reduced with the increase of the automation's control authority $\lambda_A$. However, the conventional driver model requires more control effort than the proposed model as it does not consider the controller assistance when steering. This result validates the proposed driver model. Intuitively, in path following tasks, drivers would be less engaged in vehicle control if they are aware of the controller assistance.

\subsection{Obstacle Avoidance} \label{sec.OA}
\begin{figure}[thpb]
      \centering
      \subfigure[]{
      \includegraphics[scale=0.5]{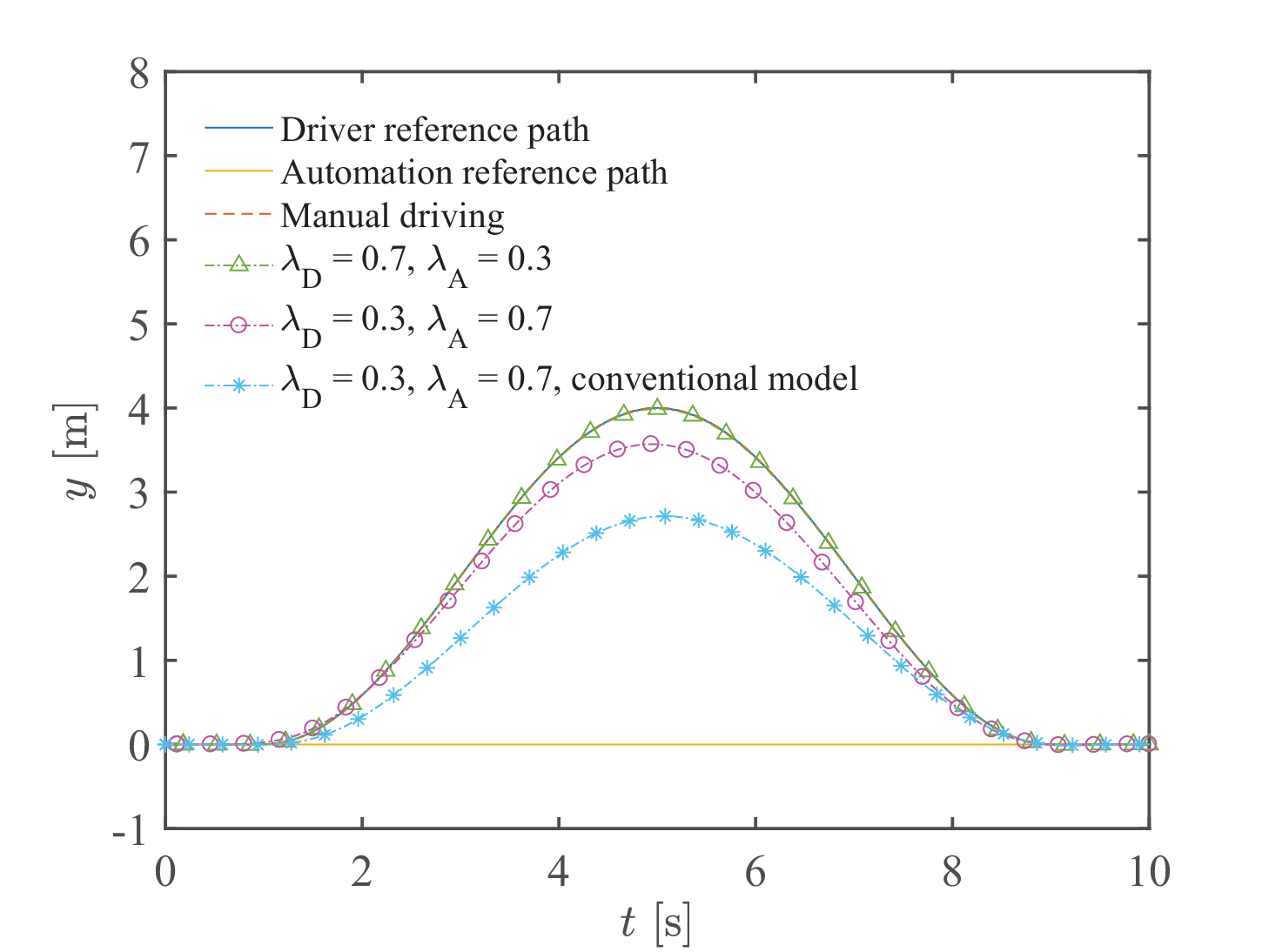}
      \label{fig.OA_y}}
      \subfigure[]{
      \includegraphics[scale=0.5]{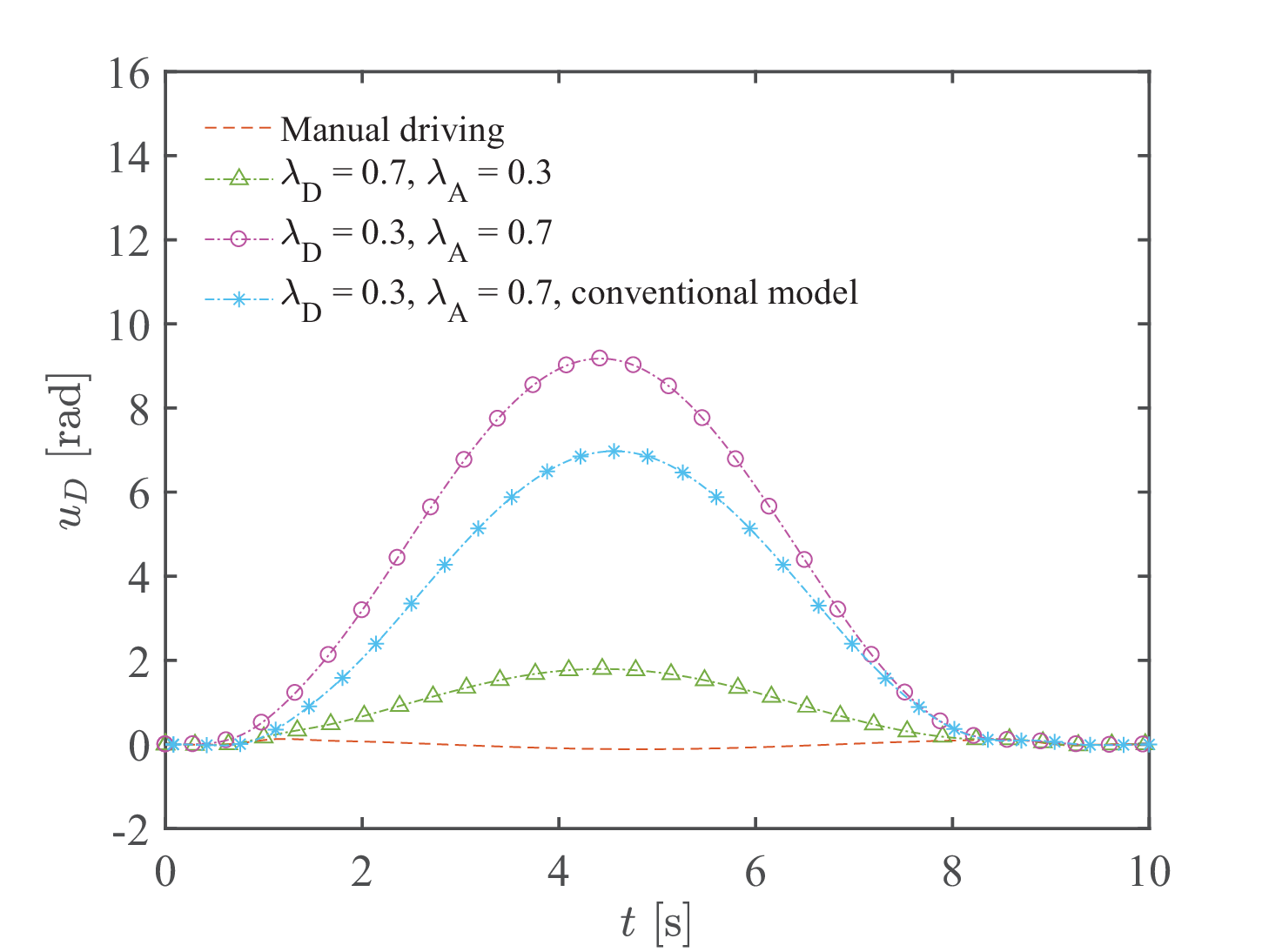}
      \label{fig.OA_u_D}}
      \caption{Obstacle-avoidance scenario: a) vehicle lateral displacement; b) driver's control effort. }
      \label{fig.OA}
\end{figure}

This part investigates how shared control reacts when the driver intention is inconsistent with the automation. In particular, we simulate the case of an obstacle undetected by the automation. Figs. \ref{fig.OA_y} and \ref{fig.OA_u_D} illustrate the vehicle lateral displacement $y$ and the driver control effort $u_D$, respectively. In this scenario, the driver wants to avoid the obstacle, whereas the automation is unaware of it and still sticks to the original reference path. We see in Fig. \ref{fig.OA_y} that a larger $\lambda_A$ impairs the obstacle-avoidance performance, as the driver's reference path is more difficult to track. This is because the controller attempts to tug the vehicle back to its reference path when the vehicle deviates. Fig. \ref{fig.OA_u_D} shows that the driver needs more effort to compensate for the controller's tug with a larger $\lambda_A$. Therefore, a larger $\lambda_A$ is undesired when the driver intentionally departs from the original reference path. Fig. \ref{fig.OA_y} also shows that if the driver does not take the controller behavior into account as in the conventional model, they will not require more control effort to compensate for the controller's tug. Consequently, the obstacle-avoidance performance further deteriorates. This however contradicts to our intuition as drivers are expected to override the automation's ``malicious" effort if they are well aware of it. Again, the proposed driver model is validated through this scenario.

\subsection{Automatic Weight Switching}
The above simulation results show that the control authority weights should vary with respect to the specific situation. If the driver intention is consistent with the automation, a larger $\lambda_A$ (smaller $\lambda_D$) will help relieve the driver's control effort and enhance path-tracking performance; otherwise a smaller $\lambda_A$ (larger $\lambda_D$) is desirable to allocate more authority to the driver. A possible strategy to implement this is given by the automatic weight switching method proposed in Section \ref{sec.weight_switching}. This part verifies the effectiveness of the automatic weight switching algorithm (\ref{eq.incons_detection}) in a complex scenario. The complex scenario is exactly a temporal combination of the path-following scenario in Section \ref{sec.PF} and the obstacle-avoidance scenario in Section \ref{sec.OA}. The parameters for automatic weight switching are chosen as: window length $H = 50$, threshold $\delta^* = 0.1$ rad, $\lambda_D^+ = \lambda_A^+ =0.7$, $\lambda_D^- = \lambda_A^- =0.3$. To take the driver model error into account, the estimated driver weighting matrix for path following is $Q_D = \diag(0.028, 0.015)$ compared with the actual one $Q_D = \diag(0.036, 0.02)$.

\begin{figure}[thpb]
      \centering
      \subfigure[]{
      \includegraphics[scale=0.5]{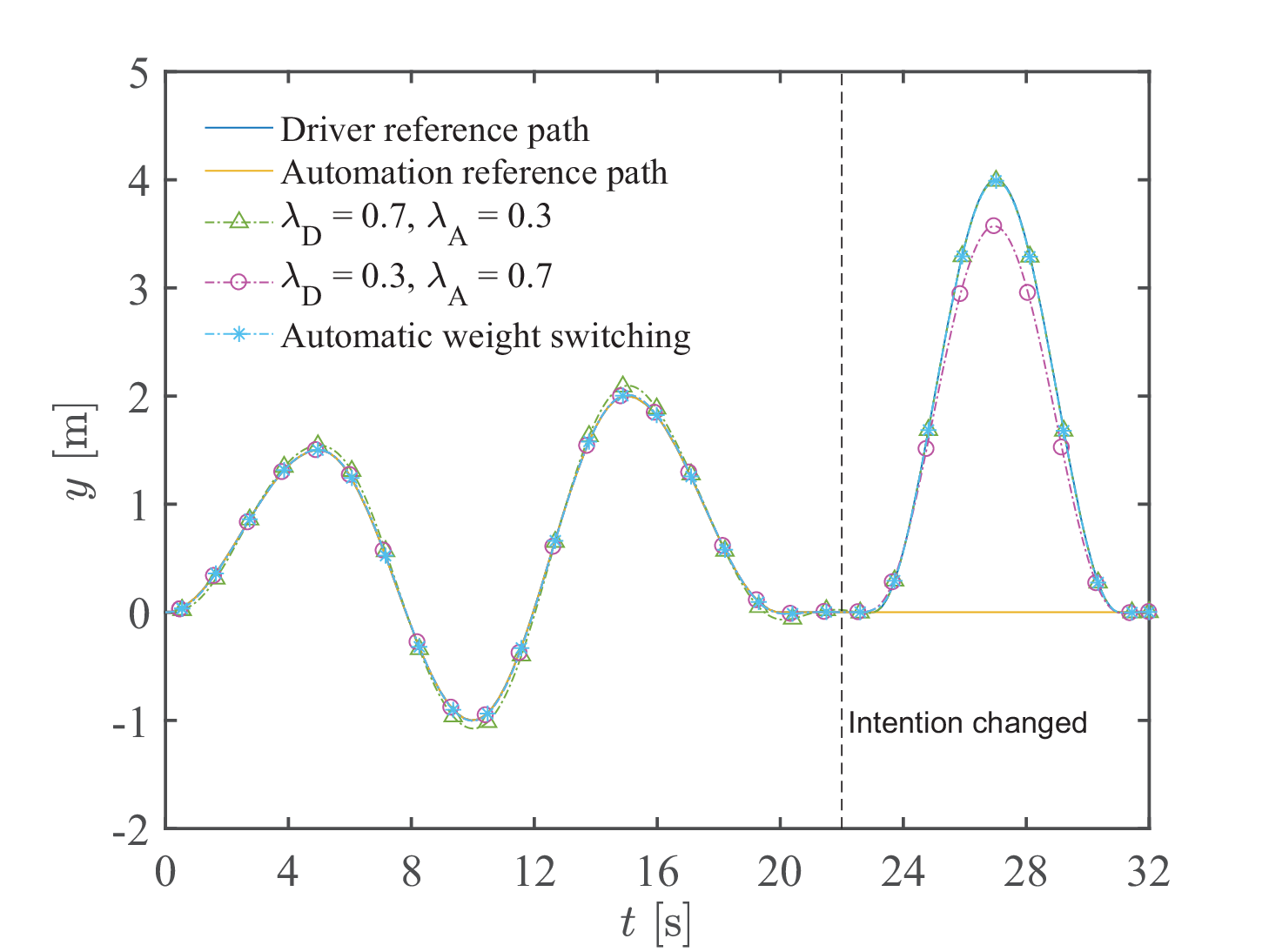}
      \label{fig.PFOA_y}}
      \subfigure[]{
      \includegraphics[scale=0.5]{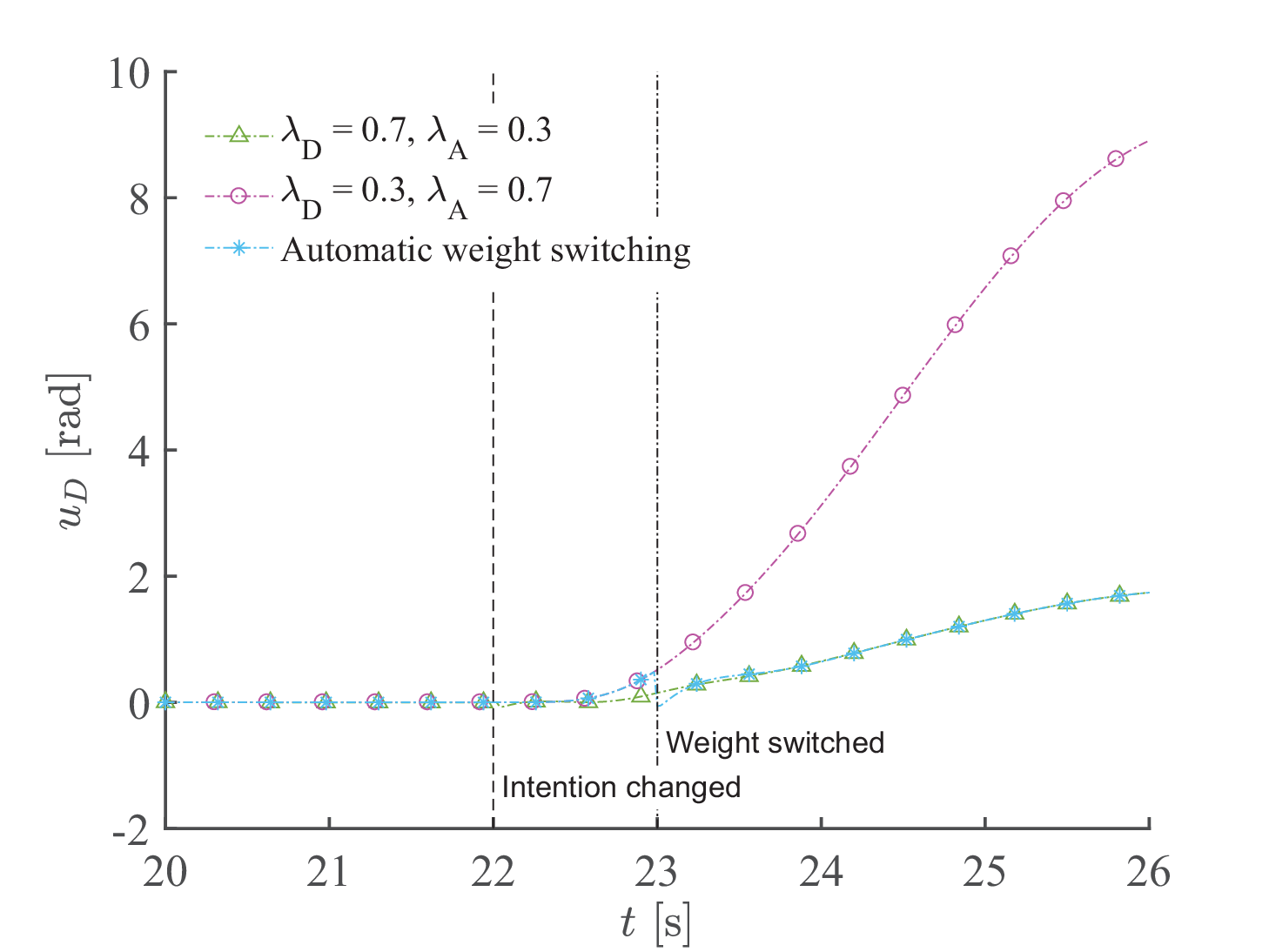}
      \label{fig.PFOA_u_D}}
      \caption{Complex scenario: a) vehicle lateral displacement; b) driver's control effort around the switching point.}
      \label{fig.PFOA}
\end{figure}

Fig. \ref{fig.PFOA} illustrates the benefits of automatic weight switching relative to static weighting. Static weighting can hardly avoid the trade-off between the performances in path following and obstacle avoidance, whereas weight switching is demonstrated to be a promising solution to it. To clearly show how the automatic weight switching works, we zoom in the $u_D$ around the switching point, as shown in Fig. \ref{fig.PFOA_u_D}. It is observed that the driver's intention change is detected 1s after it occurs, which exactly equals to the length of the sliding window. In fact, the detection time is highly correlated with the selected $\delta^*$. We can anticipate that a larger $\delta^*$ will lead to a longer detection time. Besides, we can see a drop in $u_D$ immediately after the weights are switched, which is inherently caused by the discontinuity of $\lambda_D$ and $\lambda_A$. This is a drawback of the weight switching method, which calls for future work.

\section{Conclusion} \label{sec.conclusion}
This paper investigated an indirect driver-automation shared control method for steer-by-wire vehicles. A weighted summation approach was used to balance the interests of the driver and automation, where the control authorities are defined as the weights assigned to their inputs. An MPC-based method was used to model the driver steering behavior in indirect shared control. The resulting algorithm considers the driver adaptation, assuming that they have identified the transformation strategy of the controller through motor learning, and incorporates it into the internal model for predictive control. To avoid the performance trade-off between different situations (whether the driver intention is consistent with the automation or not) stemming from static control authority allocation, this paper proposed a sliding-window-based approach to monitor the driver intention in real time, and switches the authority weights automatically. The simulation result showed that:
\begin{enumerate}
\item For path-following tasks, the driver control effort is reduced and the vehicle tracking performance is improved, with higher automation authority and lower driver authority. Compared with conventional driver model, the modeled driver becomes less engaged in control with control authority partly undertaken by the automation.
\item For obstacle-avoidance tasks, the driver has to use a larger control effort to avoid forward collision when the automation authority is higher. This shows that if the driver intention is inconsistent with the automation, indirect shared control will hinder his completion of the task. Compared with conventional driver model, the proposed model suggests that the driver manages to avoid the obstacle more smoothly by using a larger control effort as he has included the controller impedance into his internal model.
\item The proposed automatic weight switching method can effectively address the performance trade-off caused by static weighting. However, it is observed that the driver input experiences a sudden drop after the authority weights are switched.
\end{enumerate}

Limitations of the proposed driver modeling and weight switching method include the assumption that drivers can accurately identify the transformation strategy of the controller through learning. While motor control studies have shown that humans are generally able to identify complex dynamics \cite{Burdet13MITpress}, it remains to test how well the vehicle dynamics and the controller strategy can be identified in practice. Second, the weight switching approach requires improvement in terms of robustness for different drivers and to deal with the driver input discontinuity. This entails the identification of a subject specific threshold and the development of a smooth weight shifting process.

\bibliographystyle{ieeetr}
\bibliography{ref_sharedcontrol}
\end{document}